\PassOptionsToPackage{table,xcdraw,dvipsnames}{xcolor}
\documentclass[sigconf]{acmart}
\usepackage{caption}
\usepackage{subcaption}

\usepackage{algorithm}
\usepackage{algorithmicx}
\usepackage[noend]{algpseudocode}

\newcommand*\Input[1]{\Statex \textbf{Input:} #1}
\newcommand*\Output[1]{\Statex \textbf{Output:} #1}
\algrenewcommand\alglinenumber[1]{#1}

\usepackage{enumitem}
\setlist[itemize]{leftmargin=*, topsep=0pt, itemsep=0pt, parsep=0pt, partopsep=0pt}
\setlist[enumerate]{leftmargin=*, topsep=0pt, itemsep=0pt, parsep=0pt, partopsep=0pt}

\usepackage{caption}
\usepackage{subcaption}

\newtheoremstyle{mystyle}
  {\abovedisplayskip}     
  {\belowdisplayskip}     
  {}                      
  {}                      
  {\bfseries}             
  {.}                     
  {.5em}                  
  {}                      

\theoremstyle{mystyle}
\newtheorem{definition}{Definition}

\AtBeginDocument{%
  \providecommand\BibTeX{{%
    \normalfont B\kern-0.5em{\scshape i\kern-0.25em b}\kern-0.8em\TeX}}}

\copyrightyear{2021}
\acmYear{2021}
\setcopyright{acmlicensed}
\acmConference[CIKM '21]{Proceedings of the 30th ACM International Conference on Information and Knowledge Management}{November 1--5, 2021}{Virtual Event, Australia}
\acmBooktitle{Proceedings of the 30th ACM International Conference on Information and Knowledge Management (CIKM '21), November 1--5, 2021, Virtual Event,  Australia}
\acmPrice{15.00}
\acmDOI{10.1145/3459637.3481942}
\acmISBN{978-1-4503-8446-9/21/11}

\begin{document}

\fancyhead{}
\title{Online Multi-horizon Transaction Metric Estimation with Multi-modal Learning in Payment Networks}
\author{Chin-Chia Michael Yeh, Zhongfang Zhuang, Junpeng Wang, Yan Zheng, \\
Javid Ebrahimi, Ryan Mercer$^*$, Liang Wang, and Wei Zhang}
\affiliation{%
  \institution{Visa Research, University of California, Riverside$^*$}
}
\email{{miyeh, zzhuang, junpenwa, yazheng, jebrahim, liawang, wzhan}@visa.com, rmerc002@ucr.edu}

\begin{abstract}
Predicting metrics associated with entities' transnational behavior within payment processing networks is essential for system monitoring.
Multivariate time series, aggregated from the past transaction history, can provide valuable insights for such prediction.
The general multivariate time series prediction problem has been well studied and applied across several domains, including manufacturing, medical, and entomology.
However, new domain-related challenges associated with the data such as \textit{concept drift} and \textit{multi-modality} have surfaced in addition to the \textit{real-time requirements} of handling the payment transaction data at scale.
In this work, we study the problem of multivariate time series prediction for estimating transaction metrics associated with entities in the payment transaction database.
We propose a model with five unique components to estimate the transaction metrics from multi-modality data.
Four of these components capture \textit{interaction}, \textit{temporal}, \textit{scale}, and \textit{shape} perspectives, and the fifth component
fuses these perspectives together.
We also propose a hybrid offline/online training scheme to address concept drift in the data and fulfill the real-time requirements.
Combining the estimation model with a graphical user interface, the prototype transaction metric estimation system has demonstrated its potential benefit as a tool for improving a payment processing company's system monitoring capability.
\end{abstract}


\maketitle

\section{Introduction}
\label{sec-introduction}
The main responsibility of a payment network company like Visa is facilitating transactions between different parties with reliability, convenience, and security.
For this reason, it is crucial for a payment network company to actively study the transaction behavior of different entities (e.g., countries, card issuers, and merchants) in real-time because any unexpected disruption could impair users' experience and trust towards the network.
For example, an unexpected network outage could be detected promptly if a payment network company monitors the transaction volumes\footnote{Transaction volume, decline rate, and fraud rate are standard transaction metrics used in payment network companies.} of different geographical regions~\cite{mora2019predicting}.
Fraudulent activities, such as the infamous ATM cash-out attack~\cite{arun2020atm}, could be easily spotted if the abnormal decline rate$^1$ of the attacked card issuer is monitored.
Additionally, it is possible to enhance the existing fraud detection system if the real-time fraud rate$^1$ of the associated merchant is estimated and supplied to the fraud detection system~\cite{zhang2020transaction}.
Other business services provided by payment network companies, like ATM management~\cite{pearson2021real} and merchant attrition~\cite{kapur2020solving}, can also benefit from an effective monitoring system.

For monitoring purposes, we propose a real-time transaction metric estimation system that estimates the future transaction metrics of a given set of entities based on their historical transaction behaviors.
Such estimation capability can help payment network companies' monitoring effort in two ways: 1) we can detect abnormal events (e.g., payment network outage, ATM cash-out attack) by comparing the estimated transaction metric with the real observation, and introduce necessary interventions in time to prevent further loss, and 2) the online estimation provides instant access to transaction metrics (e.g., fraud rate$^1$) that are not immediately available due to the latency introduced from data processing, case investigation, and data verification.

We face the following three challenges when building the prototype transaction metric estimation system:

\textbf{Challenge 1: Concept Drift.}
In payment networks, the transaction behaviors of entities are always evolving.
A static prediction system will always become outdated when a sufficient amount of time has passed because of such incremental concept drift~\cite{gama2014survey}.
Moreover, external factors like economy, geopolitics, and pandemic could abruptly change the transaction behaviors, subsequently causes a sudden concept drift~\cite{gama2014survey} and degrades the quality of the prediction made by a static prediction system.
Concept drift is an added layer of complexity in real-world production data often absent from cleaned experimental datasets.

\begin{figure}[ht]
    \centering
    \begin{subfigure}[b]{0.45\textwidth}
        \centering
        \includegraphics[width=\textwidth]{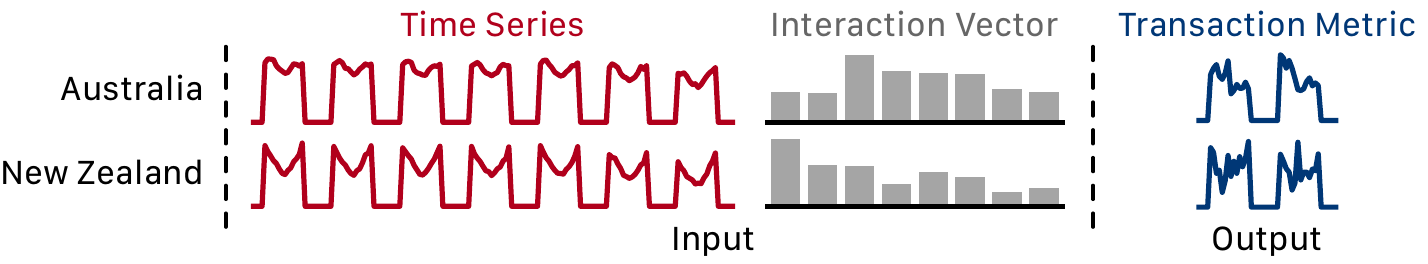}
        \caption{Australia and New Zealand have similar time series features.
        However, due to the differences in interaction vector, the transaction metric prediction is different.}
        \label{fig:vis_input_ts_au_nz}
    \end{subfigure}
    \begin{subfigure}[b]{0.45\textwidth}
        \centering
        \includegraphics[width=\textwidth]{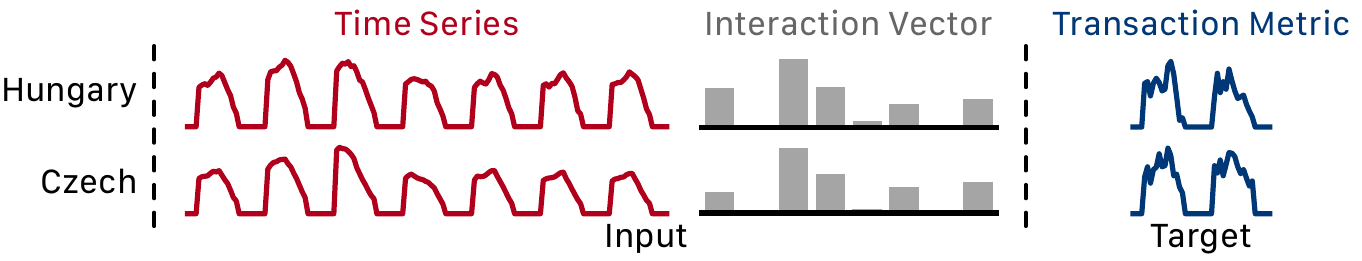}
        \caption{Hungary and Czech have similar time series features and interaction vectors.
        As a result, the transaction metric prediction is similar as well.}
        \label{fig:vis_input_ts_hu_cz}
    \end{subfigure}
    \caption{Time series features and interaction vector feature are both essential for predicting transaction metric. }
    \label{fig:vis_input_ts}
\end{figure}

\textbf{Challenge 2: Multiple Modalities.}
Because many of the aforementioned applications require our system to estimate time-varying patterns within transaction metrics, we use multivariate time series to represent the dynamic aspect of transaction behaviors.
However, recent work studying merchant behaviors using time series has demonstrated that time-series alone is \textit{not} sufficient for making accurate predictions~\cite{yeh2020merchant}.
Instead, a multi-modality approach that incorporates the relationship or \textit{interaction} among merchants is preferred.
Similar phenomena can be observed with other types of entities in the payment network.
As shown in Figure~\ref{fig:vis_input_ts_au_nz}, the time series of Australia and New Zealand exhibit similar trends, but due to the difference between their interaction vectors, their corresponding target transaction metrics have different patterns.
For another country-pair of Hungary and the Czech Republic (see Figure~\ref{fig:vis_input_ts_hu_cz}), since both time series and interaction vectors show similar patterns, the transaction metrics are identical too.
In other words, additional modalities further distinguish countries with similar time series, and the system could use such information to better estimate the target country-specific transaction metric.
Nevertheless, conventional approaches~\cite{taieb2015bias,wen2017multi,shih2019temporal,fan2019multi,zhuang2020multi,yeh2020multi} lack the ability and means to utilize the additional modalities.

\textbf{Challenge 3: Multiple Time-Step Predictions.}
With a significant amount of transactions in a system, running predictions for each time step is unrealistic due to the prohibitively high operation cost.
A more realistic approach would be to predict multiple time-steps at once.
This problem is known as \textit{multi-horizon} time series prediction, and it requires a unique model design to handle such a situation.
Merely applying a 1-step prediction model in a rolling fashion could lead to inferior results as the prediction for a later time-step is made based on estimated input.
That is, errors from the earlier time-step would propagate to later time-steps~\cite{taieb2015bias,wen2017multi}.

We use \textit{online learning} techniques to resolve challenges 1.
Online learning is a supervised learning scheme that the training data is made available \textit{incrementally} over time.
Several notable works in online learning~\cite{vzliobaite2010learning,gama2014survey,krawczyk2017ensemble,hoi2018online,losing2018incremental,lu2018learning} have been published in recent years.
Although the concept drift challenge is studied in these works, it is not well-tailored for transaction metric estimation using deep learning models.
As a result, we have outlined and benchmarked a set of online learning strategies suitable for deep learning models and identify the best strategy for our prototype.

To tackle challenges 2 and 3, we design a deep learning model capable of estimating multi-horizon transaction metrics using multiple modalities.
The proposed model has five unique components: interaction encoder, temporal encoder, scale decoder, shape decoder, and amalgamate layer.
The interaction encoder is used to process the interaction modality, i.e., how entities interact with each other in our data.
The temporal encoder dissects the temporal data and learns the inherent patterns.
The scale and shape decoders provide two distinct yet related perspectives regarding the estimated multi-horizon transaction metric.
Finally, an amalgamate layer fuses the outputs of scale and shape decoder to synthesize the output.

Our \textbf{contributions} include:
\begin{itemize}
    \item We propose and benchmark several deep learning-friendly online learning schemes to incrementally learn from transaction data.
    These online learning schemes address the concept drift challenge and allow us to avoid the expensive processes of manually re-training and re-deploying our deep learning model.
    \item We design a multi-modal predictive model that utilizes the temporal information and the interaction among different entities.
    \item We design a multi-horizon predictive model that synthesizes the target transaction metric by predicting the transaction data patterns from different perspectives.
    \item Our experiment results on real-world data have demonstrated the effectiveness of our prototype system.
\end{itemize}
\section{Notations and Problem Definition}
In this work, we specifically focus on estimating the hourly \textit{target transaction metric} for cross-border transactions based on each card's issuing country.
As a result, each entity in our prototype is a \textit{country}.
We first present how an \textit{entity} from transaction data is represented in the proposed estimation system.
An \textit{entity} is represented with two alternative views: \textit{entity time series} and \textit{entity interaction vector}.
\begin{definition}[Entity Time Series]
    {\rm
    An \textit{entity time series} of an entity $E$ is a multivariate time series~${\mathbf{T}}_\text{E} \in \mathbb{R}^{\tau_e \times d}$, where $\tau_e$ is the length of the time series and $d$ is the number of features.
    We use ${\mathbf{T}}_\text{E} [i:j]$ to denote the subsequence starting at the $i$-th timestamp and end at the $j$-th timestamp.
    }
    \label{def-ets}
\end{definition}

For example, the $\tau_e$ for the entity time series in our training data is $8,760$ (i.e., $24 \times 365$ for the entire year of 2017). The $d$ value is $14$ representing the number of features being used.

\begin{definition}[Entity Interaction Vector]
    {\rm
    Given an entity $E$ within an entity set $\mathbb{E}$ containing $k$ entities, the \textit{entity interaction vector} $\mathbf{I}_\text{E} \in \mathbb{R}^k$ is defined as the amount of interaction between $E$ and each of the $k$ entities.
    Note, $\mathbf{I}_\text{E}$ varies over time, and we use $\mathbf{I}_\text{E}[i]$ to denote the interaction vector at $i$-th timestamp.
    }
    \label{def-eiv}
\end{definition}

In our system, the entity interaction vector is a vector of size $233$ since there are $233$ country and region codes in our system.
Each element within each entity interaction vector represents the number of transactions made by cards issued in one country being used in another country within the past $30$ days.
The entity interaction vectors are generated every day.

\begin{figure}[ht]
    \centering
    \includegraphics[width=0.9\linewidth]{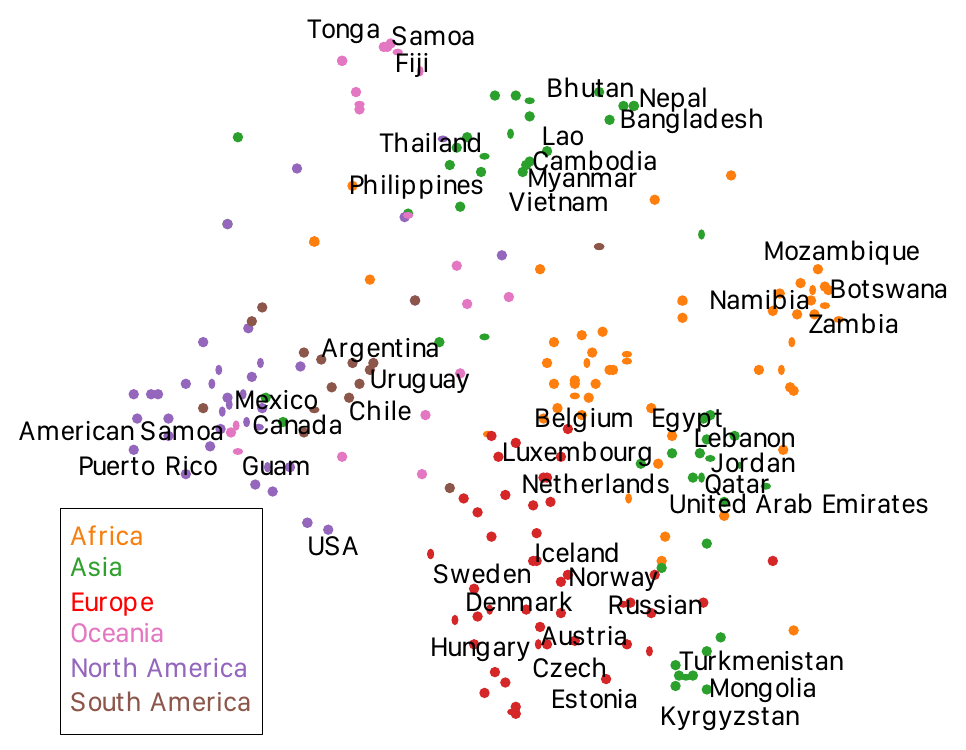}
    \caption{The interaction vectors capture the geographic and political relationship between countries.}
    \label{fig:vis_input}
\end{figure}

Because we can observe country-level patterns in our real-world datasets, it is not an arbitrary decision to use \textit{country code} as the entity.
As an example, we project each country's interaction vector from January 2017 to a two-dimensional space using $t$\texttt{-SNE}~\cite{maaten2008visualizing}, and plot the projected space in Figure~\ref{fig:vis_input}.
The interaction vectors encapsulate the similarities and dissimilarities among different countries.

\begin{definition}[Entity Transaction Metric Sequence]
    {\rm
    An \textit{entity transaction metric sequence} is a time series, denoted as $\textbf{M}_\text{E} \in \mathbb{R}^{\tau_m}$ where $\tau_m$ is the length of the time series.
    We use $\textbf{M}_\text{E}[i:j]$ to denote the subsequence starting at the $i$-th timestamp and end at the $j$-th timestamp.
    }
    \label{def-efrs}
\end{definition}

The entity time series $\mathbf{T}_\text{E}$ and entity transaction metric sequence $\textbf{M}_\text{E}$ have the identical length (i.e., $\tau_e = \tau_m$) since $\mathbf{T}_\text{E}$ and $\textbf{M}_\text{E}$ have the same sampling rate in our system.
Specifically, $\textbf{M}_\text{E}$ stores each country's target hourly transaction metric, and we have $\tau_m=8,760$ for a year with 365 days.
Although the target metric in the implemented system is a \textit{uni}-variate time series, it is possible to extend the proposed approach to \textit{multi}-variate metric sequence with techniques introduced in~\cite{yeh2020multi}.
Based on above definitions, the problem that our system solves is:
\begin{definition}[]
    {\rm
    Given a time step $i$, an entity $E$, the entity interaction vector $\mathbf{I}_\text{E}[i]$, and the entity time series $\mathbf{T}_\text{E}[i - t_p : i]$, the goal of the \textit{transaction metric estimation problem} is to learn a model~$\mathsf{F}$ that can be used to predict the transaction metric of the entity between time $i - t_a$ and $i + t_b$, where $t_a$ and $t_b$ are the number of backward and forward time-steps to estimate, respectively.
    The model~$\mathsf{F}$ can be formulated as follows:
    $\hat{\textbf{M}}_\text{E}[i - t_a :i + t_b] \gets \mathsf{F}\left(\mathbf{T}_\text{E}[i-t_p:i], \mathbf{I}_\text{E}[i]\right)$
    where $\hat{\textbf{M}}_\text{E}[i - t_a :i + t_b]$ is the estimated transaction metric from $i - t_a$ to $i + t_b$.
    }
\label{def-problem}
\end{definition}

\begin{figure}[ht]
    \centering
    \includegraphics[width=0.6\linewidth]{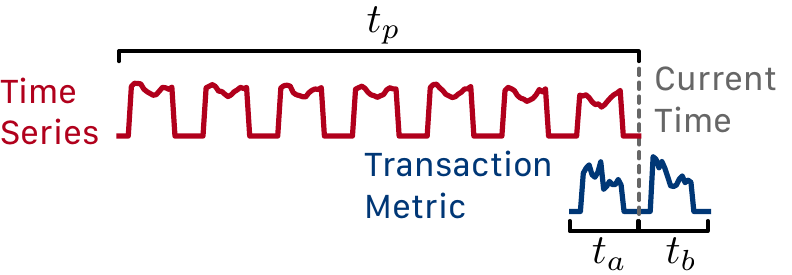}
    \caption{Relationship between the current time with various time windows (i.e., $t_p$, $t_a$, and $t_b$).}
    \label{fig:input_exp}
\end{figure}

As suggested by the notation~$\hat{\textbf{M}}_\text{E}[i - t_a :i + t_b]$, the estimated transaction metric is multi-horizontal~\cite{yeh2020multi,zhuang2020multi}.
The decision to perform multi-horizon prediction is essential for our system because it creates buffer time between each consecutive prediction to ensure there is no downtime in the production environment.
Specifically, our system predicts the \textit{target metric} for the next 24 hours ($t_b=24$).
Additionally, estimating the past for some transaction metrics is also necessary because there is a lengthy delay before the system can obtain the actual metrics.
The estimation system can generate a more accurate estimation for analysis by observing the entities' latest transaction behavior before the true target metric becomes available. 
Our system estimates the target transaction metric for the past 24 hours ($t_a=24$).
For the input entity time series, we only look back $t_p$ time-step instead of using all the available time series for better efficiency.
In our implementation, we look back 168 hours (i.e., $t_p=168$, seven days), and the past target metric is not part of the input to $\mathsf{F}$ because of the delay before the system could observe the actual target transaction metric is longer than seven days.
Figure~\ref{fig:input_exp} shows how different time windows (i.e., $t_p$, $t_a$, and $t_b$) are located temporally relative to the current time step.

Given a training set consisting of both the entity time series and the entity interaction vectors from timestamp $1$ to $\tau_e$ for each $E \in \mathbb{E}$, model~$\mathsf{F}$ is trained by minimizing the following loss function:
\begin{equation}
\sum_{E \in \mathbb{E}, i \in [1, \tau_e]} \texttt{loss}\left(\hat{\textbf{M}}_\text{E}\left[i - t_a :i + t_b\right], \textbf{M}_\text{E}\left[i - t_a :i + t_b\right]\right)
\end{equation}
where $\texttt{loss}()$ can be any commonly seen regression loss function, e.g. \textit{mean squared error}.
\section{Model Architecture}
\label{sec:model}
\begin{figure}[ht]
    \centering
    \includegraphics[width=0.65\linewidth]{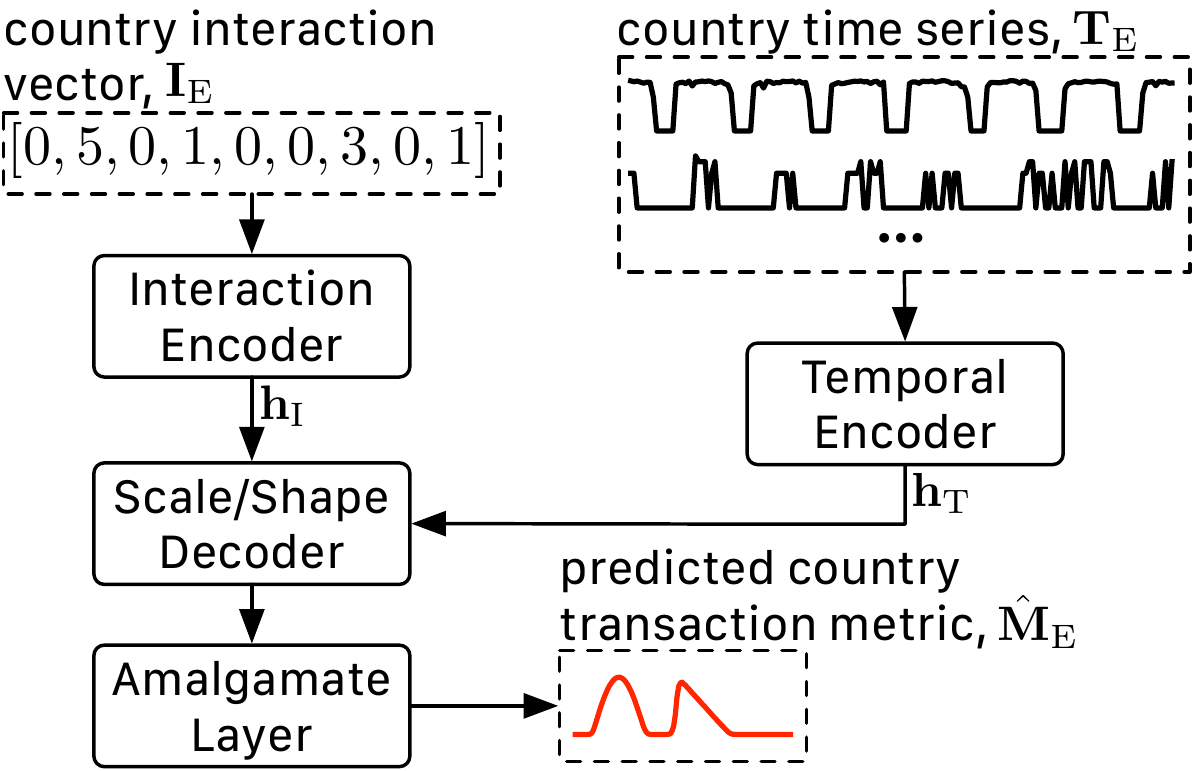}
    \caption{The transaction metric estimation model predicts the transaction metric for a period of time based on the input country's interaction vector and time series.}
    \label{fig:model_overall}
\end{figure}

Figure~\ref{fig:model_overall} depicts the overall architecture of our country-level transaction metric estimation model.
The proposed model has five components: 1) interaction encoder, 2) temporal encoder, 3) scale decoder, 4) shape decoder, and 5) amalgamate layer.
The two encoders extract hidden representations (i.e., $\mathbf{h}_\text{I}$ and $\mathbf{h}_\text{T}$) from the inputs (i.e., $\mathbf{I}_\text{E}$ and $\mathbf{T}_\text{E}$) independently.
That is, each encoder is only responsible for one aspect of the input entity.
Using the two extracted hidden representations, each decoder then independently provides information regarding different aspects of the estimated transaction metric sequence.
The scale decoder provides the scale information (i.e., magnitude and offset), and the shape decoder provides the shape information.
Lastly, the amalgamate layer combines the shape and scale information to generate the estimated transaction metric sequence.

\begin{figure}[ht]
    \centering
    \includegraphics[width=0.65\linewidth]{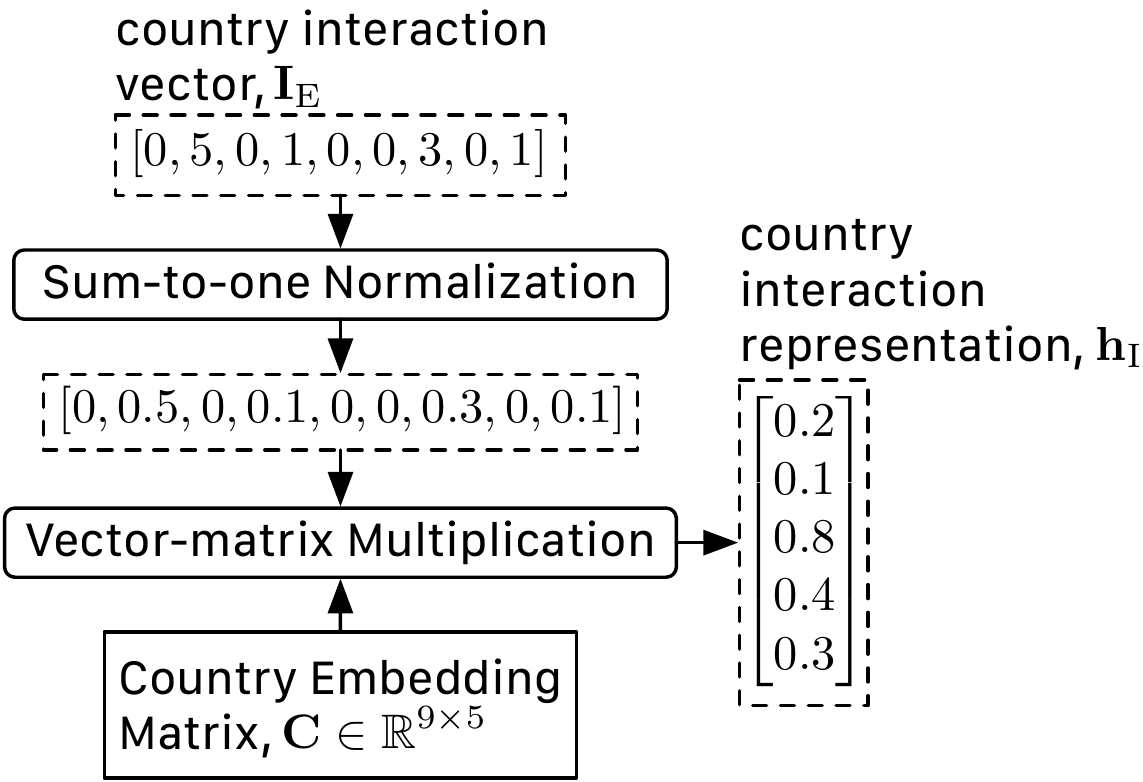}
    \caption{The design of the interaction encoder.
    Here, for demo purpose, we define the number of country as 9 and the embedding size as 5.}
    \label{fig:inter_encoder}
\end{figure}

\noindent \textbf{[Interaction Encoder]} is responsible for processing the entity interaction vectors~$\mathbf{I}_\text{E} \in \mathbb{R}^k$ where $k$ is the number of entities in the model.
The output interaction hidden representation~$\mathbf{h}_\text{I} \in \mathbb{R}^{n_k}$ captures the information about the interactions between different entities in the database where $n_k$ is the embedding size for each entity.
Following~\cite{yeh2020merchant}, we use a simple embedding-based model design shown in Figure~\ref{fig:inter_encoder}.

The input interaction vector is first normalized by the sum-to-one normalization.
This step is important for modeling interaction between different entities in transaction database~\cite{yeh2020merchant}.
Because the $L1$ vector norm of the input interaction vector is proportional to each country's population, sum-to-one normalization ensures that the hidden representation focus on capturing the information regarding the interaction between different countries instead of the population difference.

With the normalized country interaction vector~$\bar{\mathbf{I}}_\text{E}$, the interaction hidden representation~$\mathbf{h}_\text{I}$ is computed with $\bar{\mathbf{I}}_\text{E}\textbf{C}$.
The matrix~$\textbf{C}$ here contains the embeddings corresponding to each country.
In other words, the interaction hidden representation~$\mathbf{h}_\text{I}$ is a weighted sum of the embeddings in $\textbf{C}$.
The matrix~$\textbf{C}$ is the only learnable parameter in the interaction encoder.
It can either be initialized randomly or using existing embedding learned with the technique proposed in~\cite{yeh2020towards}.
The size of the returned interaction hidden representation~$\mathbf{h}_\text{I}$ depends on each country's embedding vector size.
In our implementation, the number of countries is 233, and the size of the embedding vector is set to 64.
Note, although we only model one type of interaction among the entities in our prototype, we can extend our system to handle multiple interaction types by using more interaction encoders.

\begin{figure}[ht]
    \centering
    \includegraphics[width=0.80\linewidth]{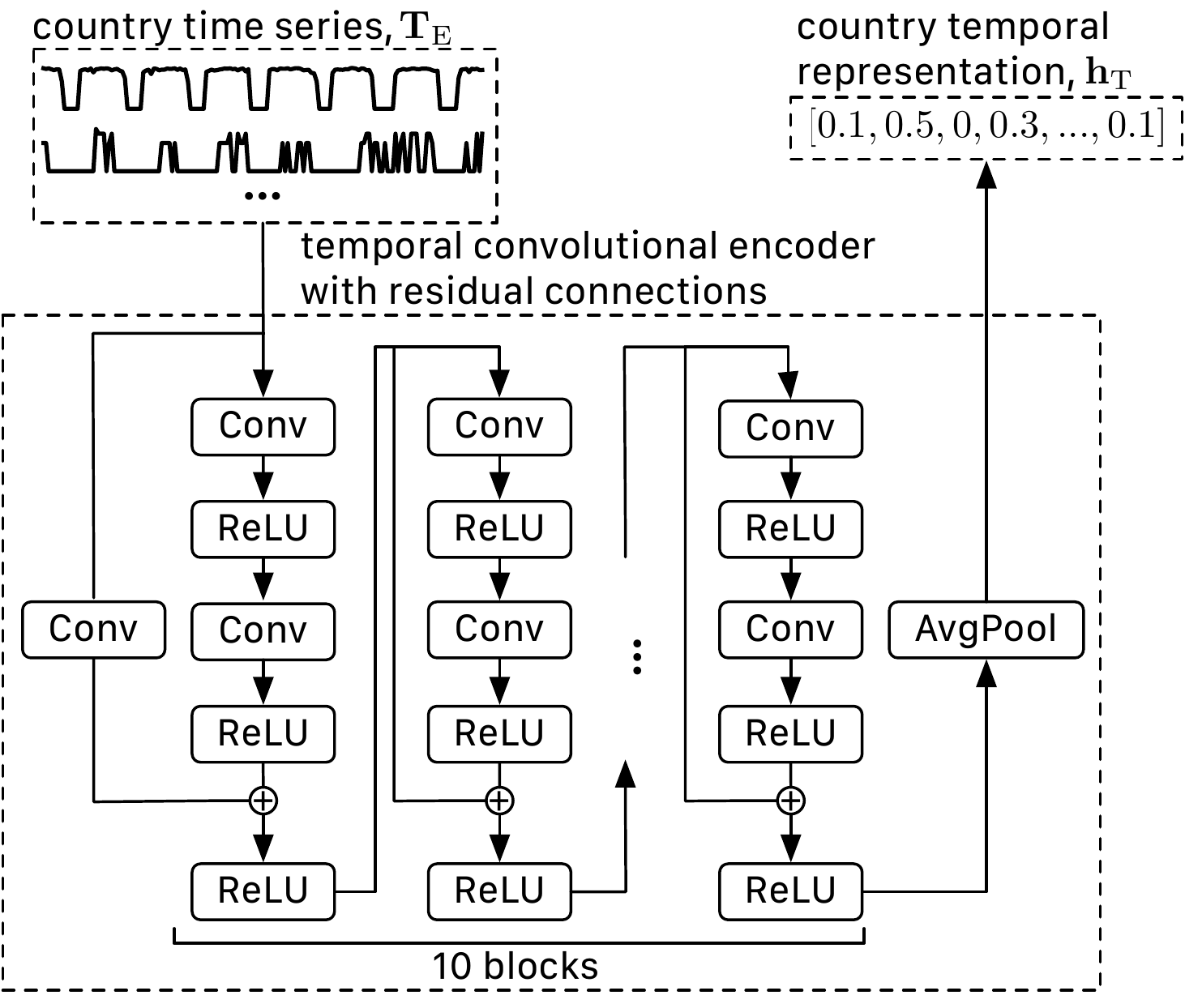}
    \caption{The design of the temporal encoder is a variant of temporal convolutional net with residual connection.}
    \label{fig:temp_encoder}
\end{figure}

\noindent \textbf{[Temporal Encoder]} extracts temporal hidden representation~$\mathbf{h}_\text{T}$ $\in \mathbb{R}^{n_k}$ from the input entity time series~$\mathbf{T}_\text{E} \in \mathbb{R}^{t_p \times d}$ where $n_k$ is the size of output hidden representation vector, $t_p$ is the length of the input time series, and $d$ is the dimensionality of the input time series.
For the particular system presented in this paper, $n_k$ is 64, $t_p$ is 168, and $d$ is 14.
As shown in Figure~\ref{fig:temp_encoder}, the temporal encoder is designed based on the Temporal Convolutional Network (TCN)~\cite{bai2018empirical} with modification suggested by~\cite{yeh2020merchant}.

The body of the model consists of a sequence of identical residual blocks.
The main passage is processed with $\texttt{Conv}\texttt{-}\texttt{ReLU}\texttt{-}\texttt{Conv}\texttt{-}\texttt{ReLU}$ then combined with the output of residual passage before passing through a $\texttt{ReLU}$ layer.
The only exception is the first residual block where it has a $\texttt{Conv}$ layer used to solve the shape mismatch problem between the output of the main passage and the residual passage.
In our implementation, all the $\texttt{Conv}$ layer has 64 kernels.
The kernel size for all the $\texttt{Conv}$ layers is 3 except the $\texttt{Conv}$ layer in the first residual passage.
The kernel size for the $\texttt{Conv}$ layer in the first residual passage is 1 as its main purpose is to match the output shape of the main passage and the residual passage.
The output temporal hidden representation is generated by summarizing the residual blocks' output across time with the global average pooling (i.e., $\texttt{AvgPool}$) layer.
Because the last $\texttt{Conv}$ layer has 64 kernels, the output temporal hidden representation~$\mathbf{h}_\text{T}$ is a vector of length 64.

\begin{figure}[ht]
    \centering
    \includegraphics[width=0.90\linewidth]{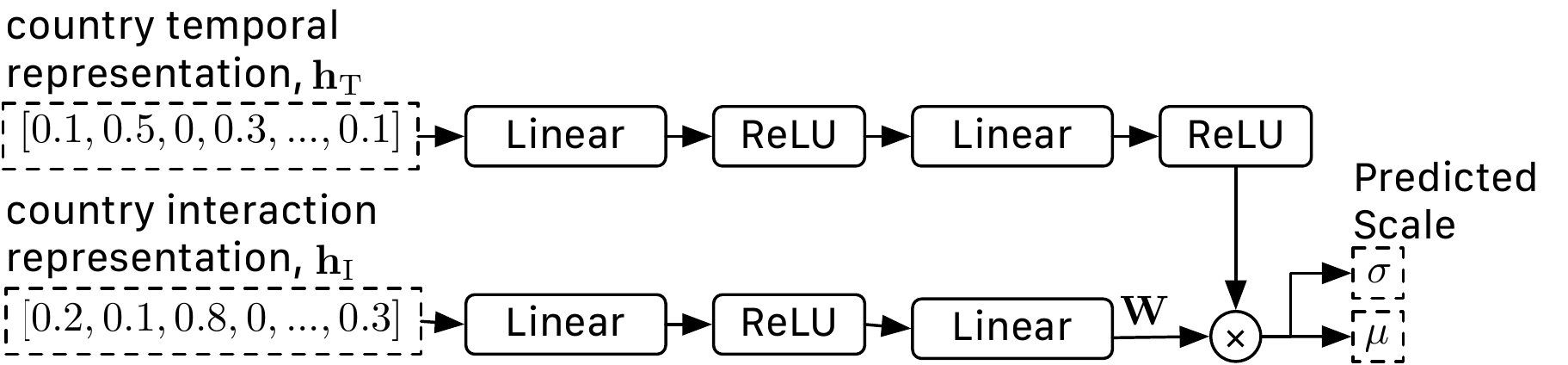}
    \caption{The scale decoder generates the magnitude~$\sigma$ and offset~$\mu$ based on the input hidden representations~$\mathbf{h}_\text{I}$ and~$\mathbf{h}_\text{T}$.}
    \label{fig:scale_decoder}
\end{figure}

\noindent \textbf{[Scale Decoder]} combines temporal hidden representation~$\mathbf{h}_\text{T}$ and interaction hidden representation~$\mathbf{h}_\text{I}$ to generate the scale (i.e., magnitude~$\sigma$ and offset~$\mu$) for the transaction metric sequence.
The structure of the scale decoder is shown in Figure~\ref{fig:scale_decoder} which extends the structure proposed in~\cite{yeh2020multi} to also consider the interaction hidden representation~$\mathbf{I}_\text{E}$ when predicting the scale.

The $\texttt{Linear}\texttt{-}\texttt{ReLU}\texttt{-}\texttt{Linear}\texttt{-}\texttt{ReLU}$ on the top further process the temporal hidden representation~$\mathbf{h}_\text{T}$ to focus on the information relevant for predicting the scale.
The $\texttt{Linear}\texttt{-}\texttt{ReLU}\texttt{-}\texttt{Linear}$ on the bottom produce a matrix~$\mathbf{W} \in \mathbb{R}^{n_k \times 2}$ which is used for mapping the output of the top $\texttt{Linear}\texttt{-}\texttt{ReLU}\texttt{-}\texttt{Linear}\texttt{-}\texttt{ReLU}$ process to~$\sigma$ and~$\mu$.
The motivation behind the design is that we assume entities with similar interaction representations should use similar $\mathbf{W}$ to estimate the scale of the target transaction metric sequence.

\begin{figure}[ht]
    \centering
    \includegraphics[width=0.99\linewidth]{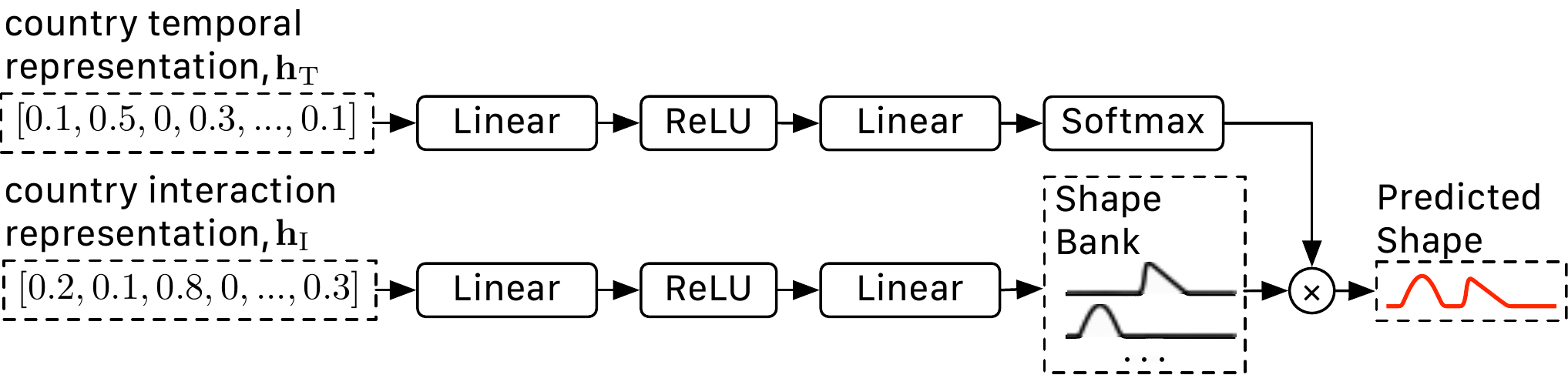}
    \caption{The shape decoder focuses on predicting the shape using the input hidden representations~$\mathbf{h}_\text{I}$ and~$\mathbf{h}_\text{T}$.}
    \label{fig:shape_decoder}
\end{figure}

\noindent \textbf{[Shape Decoder]} is responsible for providing the shape estimation for the predicted transaction metric sequence.
It combines temporal hidden representation~$\mathbf{h}_\text{T}$ and interaction hidden representation~$\mathbf{h}_\text{I}$ to generate the shape prediction.
The structure of shape decoder is similar to the scale decoder as shown in Figure~\ref{fig:shape_decoder}.

There are two differences between the design of shape decoder comparing to the design of scale encoder: 1) the output of the $\texttt{Linear}\texttt{-}\texttt{ReLU}\texttt{-}\texttt{Linear}$ for processing interaction hidden representation~$\mathbf{h}_\text{I}$ is a \textit{shape bank}~\cite{yeh2020multi} which stores basis shapes for estimating the shape of the target transaction metric sequence and 2) the last layer for processing the temporal hidden representation~$\mathbf{h}_\text{T}$ is a $\texttt{Softmax}$ layer instead of a $\texttt{ReLU}$ layer.
The output of the $\texttt{Softmax}$ layer dictates which and how the shape bank's basic shapes are combined to form the shape prediction.
Such a unique design is borrowed from~\cite{yeh2020multi}, which has been shown to outperform alternative methods for transaction data.
Similar to the scale decoder, we assume that entities with similar interaction representations should use similar basis shapes to estimate the shape of the target transaction metric sequence.

If we only consider the last two layers of the \texttt{Linear}\texttt{-}\texttt{ReLU}\texttt{-} \texttt{Linear}\texttt{-}\texttt{Softmax} passage for processing~$\mathbf{h}_\text{T}$, this sub-model is a softmax regression model.
Because the softmax regression model's output is positive and always sum-to-one, it forces the model to pick only the relevant basis shapes for the prediction.
As shown in the example presented in Figure~\ref{fig:shape_decoder}, the predicted shape is generated by combining mostly the first two basis shapes in the shape bank.
This design provides users a way to understand what the model learned from the data by looking at the shape bank's content.



\noindent \textbf{[Amalgamate Layer]} combines the scale and shape information with $\hat{\mathbf{M}}_\text{E}^{\text{shape}} \times \sigma + \mu$ where $\hat{\mathbf{M}}_\text{E}^{\text{shape}}$ is the predicted shape, $\sigma$ contains the magnitude information, and $\mu$ contains the offset information.
Following~\cite{yeh2020multi}, we minimize the following loss function:
$\sum_{E \in \mathbb{E}} \texttt{MSE}\left(\hat{\mathbf{M}}_\text{E}, \mathbf{M}_\text{E}\right) + \gamma \texttt{NMSE}\left(\hat{\mathbf{M}}_\text{E}^{\text{shape}}, \mathbf{M}_\text{E}\right)$
where $\texttt{MSE}()$ is a function computes the mean squared error, $\texttt{NMSE}()$ is a function computes the normalized mean squared error, $\gamma$ is a hyper-parameter to make sure the output of $\texttt{MSE}()$ and $\texttt{NMSE}()$ are in similar scale, and $\hat{\mathbf{M}}_\text{E}$ is the output of amalgamate layer,.
$\texttt{NMSE}()$ is computed by first $z$-normalizing~\cite{rakthanmanon2012searching} ground truth, then calculating the mean squared error between the $z$-normalized ground truth and~$\hat{\mathbf{M}}_\text{E}^{\text{shape}}$.

\section{Online Learning Scheme}
\label{sec:learning}
We focus our investigation on refining Stochastic Gradient Descent-based optimization methods. This family of methods performs well for handling concept drift under online learning scenarios~\cite{hoi2018online,losing2018incremental} and is widely applied in the training of deep learning models~\cite{lecun2012efficient}.
Algorithm~\ref{alg-train} shows the training scheme for our system.
The input to the algorithm is a set of targeted entities that users wish to model, and the output is the estimated transaction metric sequence generated daily.

\begin{algorithm}[htb]
    \centering
    {\footnotesize
    \caption{The Training Algorithm\label{alg-train}}
    \begin{algorithmic}[1]
        \Input{set of target entities~$\mathbb{E}$}
        \Output{estimated transaction metric sequence~$\hat{\mathbf{M}}_\text{E}$}
        \Function{\texttt{train}}{$\mathbb{E}$}
        \State $\mathsf{F} \gets \texttt{InitializeModel}()$
        \State $\mathsf{F} \gets \texttt{OfflineTraining}(\mathsf{F}, \mathbb{E})$
        \While{$\texttt{True}$}
        \If{\textbf{time to update is} $\texttt{False}$}
        \State continue
        \EndIf
        \State $\mathbb{E} \gets \texttt{GetLatestAvailableData}()$
        \For{$b \gets 0$ \textbf{to} $n_{\mathrm{iter}}$}
        \State $\mathbf{I}_{\text{E}, b}, \mathbf{T}_{\text{E}, b}, \mathbf{M}_{\text{E}, b} \gets \texttt{ GetMiniBatch}(\mathbb{E})$

        \State $\mathsf{F} \gets \texttt{UpdateModel}\left(\mathbf{I}_{\text{E}, b}, \mathbf{T}_{\text{E}, b}, \mathbf{M}_{\text{E}, b}\right)$
        \EndFor
        \State $\hat{\mathbf{M}}_\text{E} \gets \texttt{Estimation}(\mathsf{F}, \mathbb{E})$
        \State \Return{$\hat{\mathbf{M}}_\text{E}$}
        \EndWhile
        \EndFunction
    \end{algorithmic}}
\end{algorithm}

First, the estimation model is initialized in \texttt{Line~2}.
Next, in \texttt{Line~3}, the initialized model is first trained offline with all the available data.
The model goes into an online mode when entering the loop beginning at \texttt{Line~4}.
In \texttt{Line~5} and \texttt{Line~6}, the function checks if it is time to update the model.
The model is updated daily.
If it is time to update the model, the latest data is pulled from the transaction database in \texttt{Line~7}.
From \texttt{Line~8} to \texttt{Line~10}, the model is updated for $n_{\mathrm{iter}}$ iterations.
Mini-batch for training are sampled in \texttt{Line~9}.
We aim to refine this sampling step that is crucial for updating the model in an online learning setup as updating the model with ``irrelevant'' data could hurt the performance when concept drift is encountered.

In addition to the baseline \textit{uniform} sampling method that is commonly seen in training deep learning model, we have tested a total of seven different special sampling techniques for sampling mini-batch when updating the model during online training.
We group these seven different sampling techniques into two categories: temporal-based and non-temporal-based sampling methods.
The temporal-based sampling method consists of methods that sample based on the temporal location of each candidate samples.
The following three temporal-based sampling methods are tested.
\begin{itemize}
    \item \textit{Fixed Window} samples the training examples within the latest $x$ days uniformly where $x$ is a hyper-parameter for this method.
    In other words, it ignores older data.
    \item \textit{Time Decay} samples the training examples with a decaying probability as the data aging.
    The probability decays linearly in our implementation.
    \item \textit{Segmentation}~\cite{gharghabi2017matrix,gharghabi2019domain} (i.e., time series segmentation) can be used to identify a good window and time decay function in a data driven fashion.
    We have modified the method presented in~\cite{gharghabi2017matrix,gharghabi2019domain} for our problem.
    We present the modification in Section~\ref{sec:segmentation}.
\end{itemize}

For non-temporal-based methods, the temporal location of each candidate sample is not considered in the sampling process.
The following four non-temporal-based methods are explored.
\begin{itemize}
    \item \textit{Similarity} biases toward examples those are more similar to the current input time series~$\mathbf{T}_\text{E}[i-t_p : i]$, where $i$ is the current time.
    As it only looks at the~$\mathbf{T}_\text{E}$, it can only help the case where the concept drift affects~$\mathbf{T}_\text{E}$.
    We use MASS algorithm~\cite{mueen2017fastest} to compute the similarity.
    \item \textit{High Error} biases toward ``hard'' examples for the current model.
    Pushing the model to focus on hard examples is commonly seen in boosting-based ensemble methods~\cite{schapire2003boosting}, and we are testing this idea for our problem.
    \item \textit{Low Error} biases toward examples that can be predicted well based on the current model.
    Because the targeted transaction metric can be noisy sometimes, this sampling strategy could remove noisy samples since noisy samples tend to introduce large errors.
    \item \textit{Training Dynamic} based sampling method has been proven effective for classification tasks~\cite{swayamdipta2020dataset}.
    Particularly, it uses measures such as \textit{confidence} and \textit{variability} to sample data.
    We have computed the confidence and variability, mostly following the formulation presented in~\cite{swayamdipta2020dataset}.
    Because~\cite{swayamdipta2020dataset} focuses on a different objective (i.e., classification), we modify the original probability distribution-based formulation to an error-based formulation for computing the confidence and variability.
We test four sampling strategies using these two measures: 1) high confidence, 2) low confidence, 3) high variability, and 4) low variability.
\end{itemize}

With the updated model, estimating the transaction metric sequence using the latest data is generated in \texttt{Line~11} and returned in \texttt{Line~12}.
Note, as Algorithm~\ref{alg-train} is an online learning algorithm the while loop continues after the model is output in \texttt{Line~12}.

\subsection{Time Series Segmentation}
\label{sec:segmentation}
Our time series segmentation-based sampling method is modified from the \texttt{FLUSS} algorithm~\cite{gharghabi2017matrix,gharghabi2019domain}; it is built on top of the matrix profile idea~\cite{yeh2016matrix,yeh2017matrix,yeh2018time}.
The matrix profile is a way to efficiently explore the nearest neighbor relationship between subsequences of a time series~\cite{zhu2016matrix,yeh2018towards}.
Because each candidate example for our model is a subsequence of the entity time series, we adopt a matrix profile-based method for our problem.
We use Algorithm~\ref{alg-segmet} to compute the sampling probability of each candidate for each entity in the training data.
The input to the algorithm is the entity time series.
We only use the part of the time series where the ground truth transaction metric is available.

\begin{algorithm}[htb]
    \centering
    {\footnotesize
    \caption{The Modified \texttt{FLUSS} Algorithm\label{alg-segmet}}
    \begin{algorithmic}[1]
        \Input{Entity time series~$\mathbf{T}_\text{E}$}
        \Output{Sampling probability~$P$}
        \Function{\texttt{getProb}}{$\mathbf{T}_\text{E}$}
        \For{$j \gets 1$ \textbf{to} $d$}
        \State $T_{j,\texttt{MPI}} \gets \texttt{GetMatrixProfileIndex}(\mathbf{T}_{\text{E}}^{j})$
        \State $T_{j,\texttt{CAC}} \gets \texttt{GetCorrectedArcCount}(T_{j,\texttt{MPI}})$
        \State $P \gets \texttt{Add}(T_{j,\texttt{CAC}})$
        \EndFor
        \State $p_{\texttt{min}} \gets \texttt{inf}$
        \For{$j \gets \texttt{length}(P)$ \textbf{to} $1$}
        \If{$P[j] < p_{\texttt{min}}$}
        \State $p_{\texttt{min}} \gets P[j]$
        \Else
        \State $P[j] \gets p_{\texttt{min}}$
        \EndIf
        \EndFor
        \State $P \gets \frac{P}{\texttt{Sum}(P)}$
        \State \Return{$P$}
        \EndFunction
    \end{algorithmic}}
\end{algorithm}

\begin{figure}[ht]
    \centering
    \includegraphics[width=0.75\linewidth]{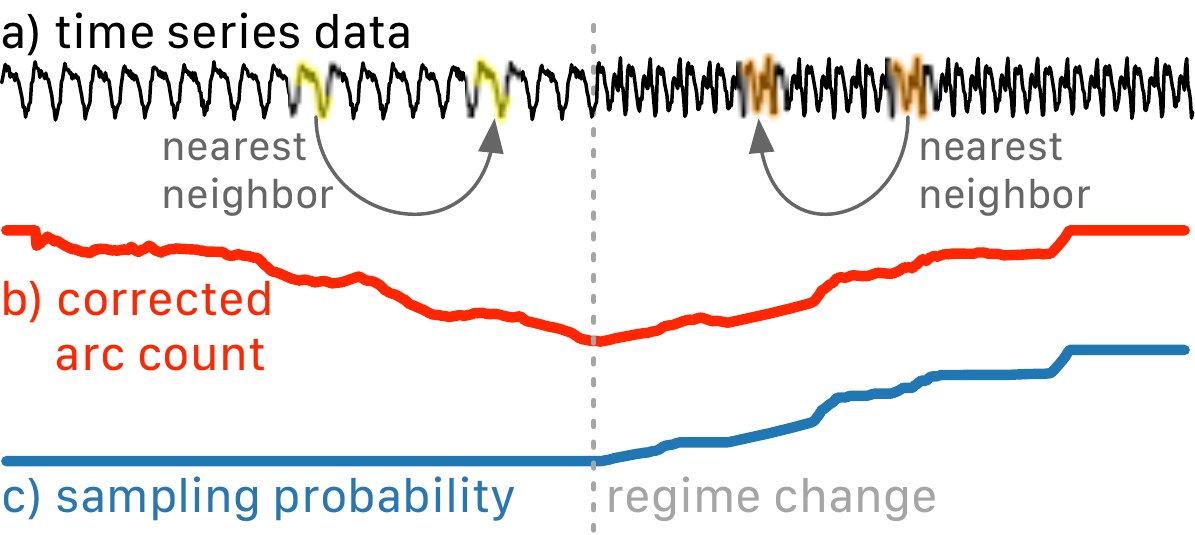}
    \caption{The sampling probability is computed from the corrected arc count curve.}
    \label{fig:segment}
\end{figure}

From \texttt{Line~2} to \texttt{Line~5}, each iteration of the loop processes each dimension of the entity time series independently.
In \texttt{Line~3}, the matrix profile index~\cite{yeh2016matrix} of the input time series's $i$-th dimension is computed.
The matrix profile index shows the nearest neighbor of each subsequence in the input time series with subsequence length $t_p$.
As shown in Figure~\ref{fig:segment}a, we can connect each subsequence with its nearest neighbor using an arc based on the information stored in the matrix profile index.

To further process the matrix profile index, we compute the corrected arc count~(CAC) curve~\cite{gharghabi2017matrix}.
The curve records the number of arcs passing through each temporal location.
We use a correction method proposed in~\cite{gharghabi2017matrix} to correct the fact that it is more likely to have an arc passing through the center relative to the two ends of the time series.
The correction method corrects the count by comparing the actual count to the expected count.
As the first half and second half of the time series data shown in Figure~\ref{fig:segment}a are different, and the arc only connects similar subsequences, there is almost no arc passing through the center of the time series, and the CAC curve is low in the center (see Figure~\ref{fig:segment}b).
To combine the CAC curve of each dimension, similar to~\cite{gharghabi2019domain}, we just add the CAC curve together in \texttt{Line~5}.

Once the CAC curve is computed, we need to convert the CAC curve to sampling probability.
In \texttt{Lines~6 -- 11}, we use the loop to enforce the non-decreasing constrain. The goal is to ensure the subsequences belonging to the newest regime (i.e., subsequences after the latest segmentation point) have higher sampling probability than those from the old regime.
As demonstrated in Figure~\ref{fig:segment}c, the non-decreasing constraint flattens the CAC curve before the regime change.
Lastly, the non-decreasing CAC curve is converted to a probability curve in \texttt{Line~12} and returned in \texttt{Line~13}.
\begin{figure}[ht]
    \centering
    \includegraphics[width=0.7\linewidth]{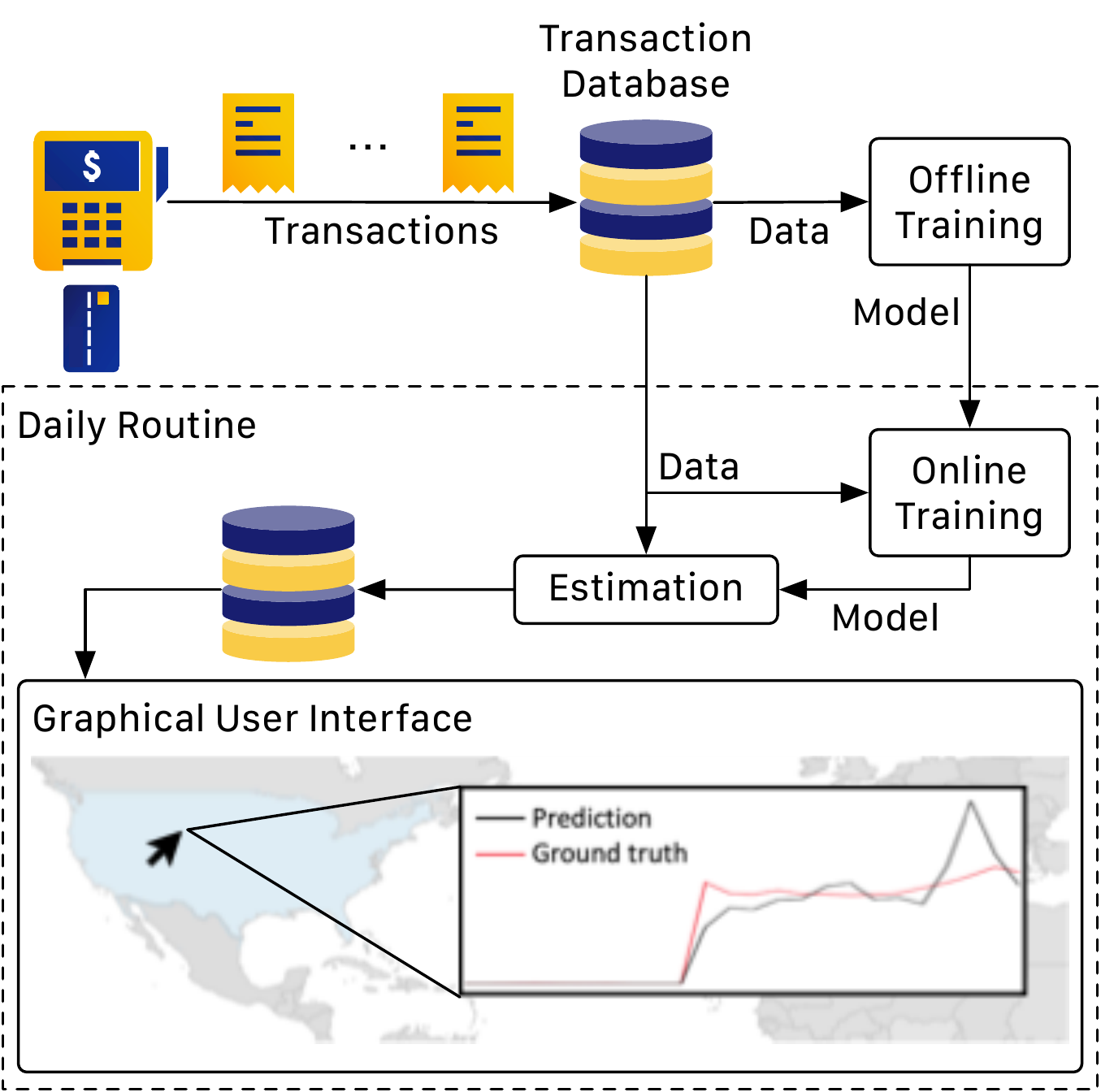}
    \caption{System design of the proposed system.
    The system consists of an offline training stage prior to deployment, a daily routine when the system is deployed, and a graphical user interface for users to monitor the system.}
    \label{fig:system}
\end{figure}

\section{System Implementation}
As presented in Section~\ref{sec:learning}, the proposed transaction metric estimation system has two training phases: the offline and the online training phases.
Before deployment, the offline training module retrieves the latest data from the transaction database to train the initial model.
Once we deploy the initial model, the online training module and estimation module would be activated daily at midnight.
The online training module first updates the current model using the best online learning scheme established in Section~\ref{sec:eval_online} with available data from the transaction database.
Next, the estimation module produces the target transaction metric estimation for the succeeding 24 hours and the preceding 24 hours.
We then stored the predictions in a database.
The user can monitor the system with a graphical user interface (GUI) by either visually evaluating the most recent prediction or comparing the previous prediction with ground truth data if the ground truth data is available.
Figure~\ref{fig:system} depicts the overall design of the system.

In the implemented prototype system, we store raw transaction data in the Hadoop Distributed File System (HDFS)~\cite{apachehadoop}.
Because of the sheer volume of transaction data, we perform most of the data processing jobs in Hadoop and generate hourly aggregated multivariate time series using Apache Hive~\cite{apachehive}.
We then move these hourly-aggregated time series to a local GPU sever for offline and online training with PyTorch~\cite{paszke2019pytorch}.
The inference/prediction is also performed on the GPU server, and the corresponding results are maintained/updated in a MySQL~\cite{mysql} database.
A web-based GUI is designed to facilitate the accessing from any client-side browsers.
An example interface of our system is shown in Figure~\ref{fig:system}, which shows that users can flexibly fetch data for any country of interest from the map visualization.
\section{Experiment}
The models and optimization process are implemented with PyTorch 1.4~\cite{paszke2019pytorch} with Adam optimizer~\cite{kingma2014adam}.
The similarity search~(i.e., MASS~\cite{mueen2017fastest}) and matrix profile~\cite{yeh2016matrix} computation are handled by Matrix Profile API~\cite{van2020mpa}.
We use country-wise aggregated transaction data from 1/1/2017 to 12/31/2017 as the training data, and data from 1/1/2018 to 12/31/2018 as the test data.
To measure the performance, we use root mean squared error~(RMSE), normalized root mean squared error~(NRMSE)~\cite{yeh2020multi,zhuang2020multi}, and coefficient of determination~($R^2$).

\subsection{Verification of the Model Architecture}
\label{sec:verification}
In this section, we evaluate different design decisions regarding our model architecture.
Notably, we test the benefit of using both entity interaction vectors and entity time series, the advantage of using the shape/scale decoder design, and comparing our best model configuration with alternative methods.
We use two different temporal encoder designs for this section: 1) the convolutional neural network~(\texttt{CNN}) which uses the design presented in Figure~\ref{fig:temp_encoder} and 2) two layers of gated recurrent units (\texttt{GRU}s)~\cite{cho2014learning}.
We only test \texttt{GRU} instead of long short-term memory~(\texttt{LSTM}) because it has been established in~\cite{yeh2020merchant,yeh2020multi,zhuang2020multi} that \texttt{GRU} has similar or superb performance comparing to \texttt{LSTM} when processing time series generated from transaction data.
In each subsequent subsection, we add a new mechanism to the base models and observe the before-and-after performance difference.

\begin{figure}
    \centering
    \begin{subfigure}[b]{0.308\linewidth}
        \centering
        \includegraphics[width=\textwidth]{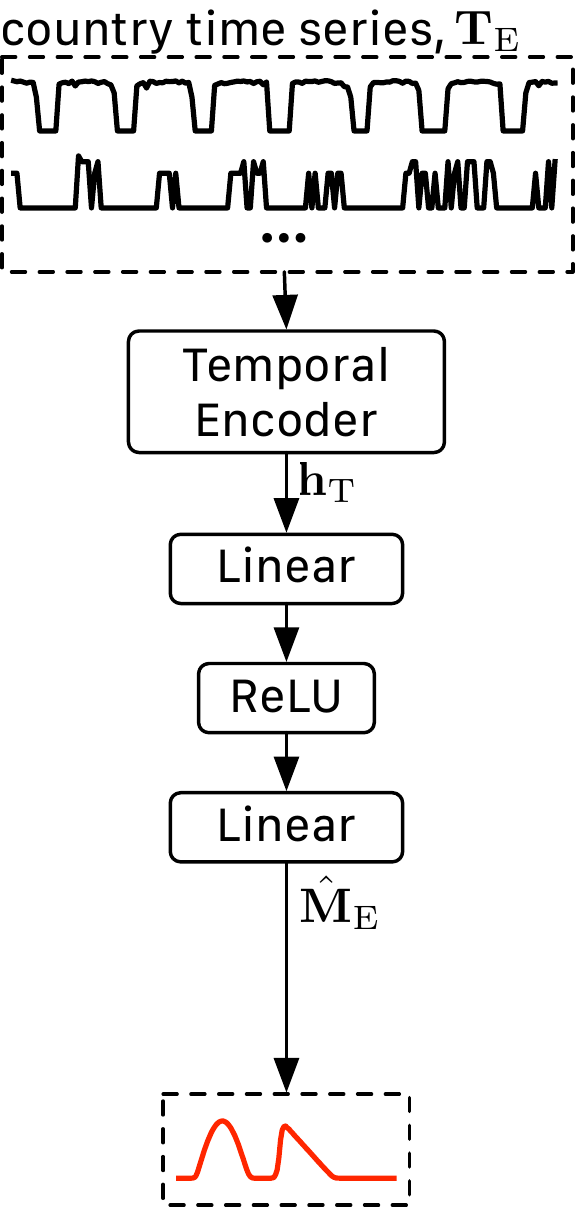}
        \caption{base model}
        \label{fig:base}
    \end{subfigure}
    \hfill
    \begin{subfigure}[b]{0.60\linewidth}
        \centering
        \includegraphics[width=\textwidth]{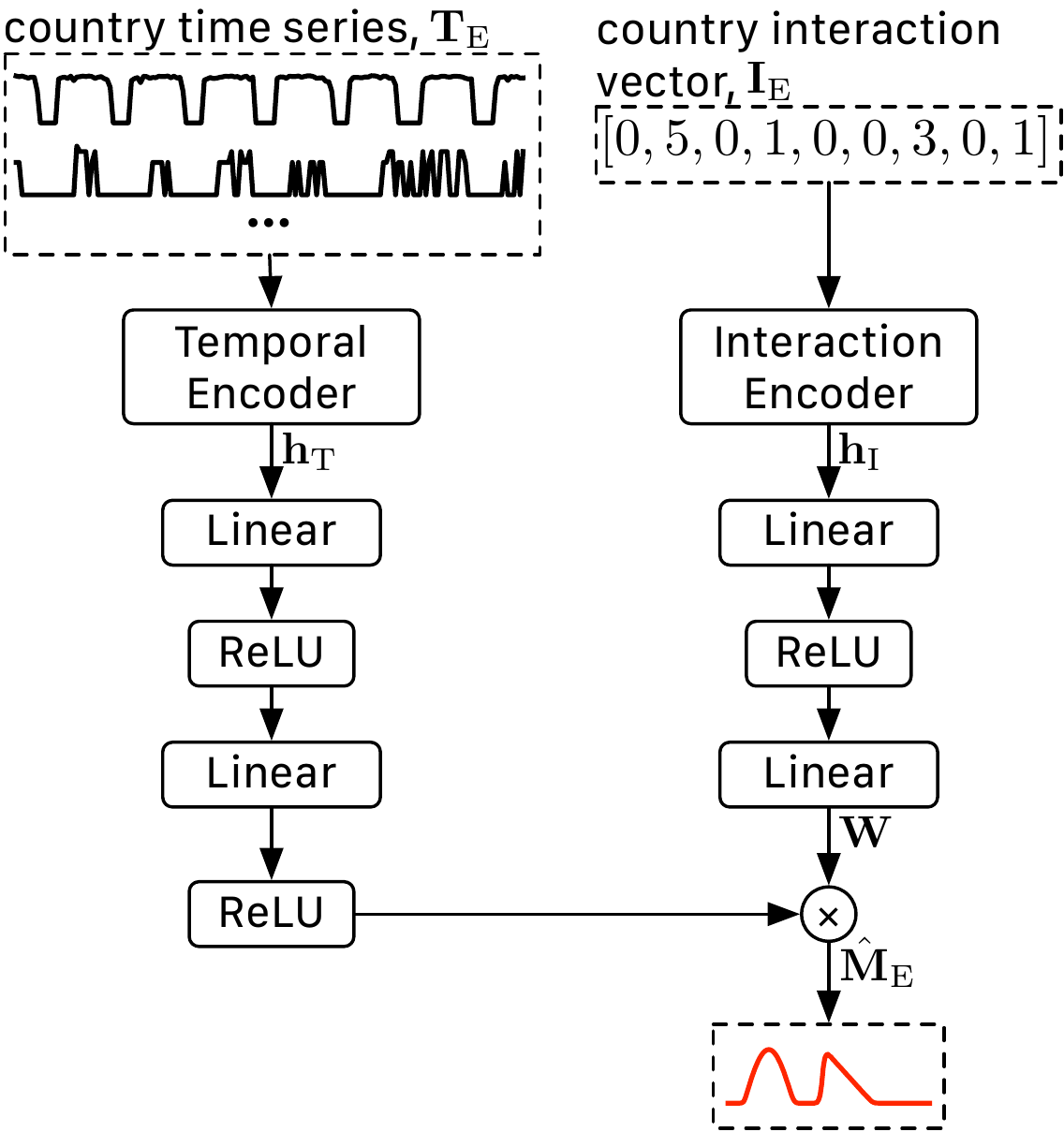}
        \caption{base model + interaction vector}
        \label{fig:base_inter}
    \end{subfigure}
    \caption{The proposed model design is compared with these baseline model designs.}
    \label{fig:alt_design}
\end{figure}

\subsubsection{Benefit of Interaction Vector}\label{sec:int_vector} \hfill \\
First, we examine the benefit of modeling both entity time series and entity interaction vector.
The base model (i.e., Figure~\ref{fig:base}) we consider only uses entity time series because it consists of richer information regarding each entity in transaction data comparing to the interaction vector as demonstrated in~\cite{yeh2020merchant}.
To evaluate the benefit of modeling both entity time series and interaction vector, we compare the base model to a more straightforward multimodality design proposed in~\cite{yeh2020merchant} (i.e., Figure~\ref{fig:base_inter}).
The model shown in Figure~\ref{fig:base_inter} is simpler comparing to the model presented in Figure~\ref{fig:model_overall} as it does not use shape/scale-decoder design.
Because neither the base model nor the multimodality model estimate the shape and scale separately, they are trained by minimizing the MSE loss function instead.
The passage processing the interaction hidden representation~$\mathbf{I}_\text{E}$ outputs a matrix~$\mathbf{W} \in \mathbb{R}^{n_k \times (t_a + t_b)}$ which is multiplied by the output of the passage process the time series hidden representation~$\mathbf{h}_\text{T}$ to generate the prediction~$\hat{\mathbf{M}}_\text{E}$.
Table~\ref{tab:base} shows the improvement of using multimodality model comparing to the base model for both \texttt{CNN}-based and \texttt{GRU}-based temporal encoder design.
Note, the base models are the baseline deep learning models.

\begin{table}[ht]
\centering
\caption{Using both time series and interaction vector improves the performance.}
\label{tab:base}
\resizebox{0.99\columnwidth}{!}{
\begin{tabular}{l|ccc|ccc}
              & \multicolumn{3}{c|}{\texttt{CNN}-based Encoder}  & \multicolumn{3}{c}{\texttt{GRU}-based Encoder}  \\
              & RMSE   & NRMSE  & $R^2$ & RMSE   & NRMSE  & $R^2$   \\ \hline
Base          & 8.2741 & 0.8545 & 0.1493 & 7.7352 & 0.8287 & 0.1943 \\
Base + Inter. & 7.3165 & 0.8256 & 0.2026 & 7.0617 & 0.8098 & 0.2268 \\  \hline
\% Improved    & 11.6\% & 3.4\%  & 35.7\% & 8.7\%  & 2.3\%  & 16.7\%
\end{tabular}
}
\end{table}

The improvement is consistent across the three different performance measurements for both temporal encoder designs.
The percent improvement is considerably more for~$R^2$ comparing to other performance measurements. The percent improvement for the models with \texttt{CNN}-based temporal encoder is more prominent than the models with \texttt{GRU}-based temporal encoder.
Overall, \texttt{GRU}-based temporal encoder performs slight better than the \texttt{CNN}-based temporal encoder.

\begin{table}[ht]
\centering
\caption{Shape/scale decoder design improves the performance.}
\label{tab:shape}
\resizebox{0.99\columnwidth}{!}{
\begin{tabular}{l|ccc|ccc}
              & \multicolumn{3}{c|}{\texttt{CNN}-based Encoder}  & \multicolumn{3}{c}{\texttt{GRU}-based Encoder}  \\
              & RMSE   & NRMSE  & $R^2$ & RMSE   & NRMSE  & $R^2$   \\ \hline
Base + Inter. & 7.3165 & 0.8256 & 0.2026 & 7.0617 & 0.8098 & 0.2268 \\
Proposed       & 5.7147 & 0.7386 & 0.2771 & 6.4736 & 0.7426 & 0.2598 \\ \hline
\% Improved    & 21.9\% & 10.5\% & 36.8\% & 8.3\%  & 8.3\%  & 14.6\%
\end{tabular}
}
\end{table}

\subsubsection{Benefit of Shape/Scale Decoder Design} \hfill\\
Next, we would like to confirm whether the shape/scale decoder design (i.e., Figure~\ref{fig:model_overall}) can further improve the performance.
We compare it to the base model shown in Figure~\ref{fig:base_inter}, which was the superior model design based on the experiment presented in Section~\ref{sec:int_vector}.
Table~\ref{tab:shape} shows how the shape/scale decoder design (i.e., the proposed design) further improves the performance.

Once again, the improvement is consistent across the three performance measurements for both temporal encoder designs.
The percent improvement is more noticeable for~$R^2$ comparing to other performance measurements.
The percent improvement for the models with \texttt{CNN}-based temporal encoder is remarkably more than the models with \texttt{GRU}-based temporal encoder.
The best model configuration according to both Table~\ref{tab:base} and Table~\ref{tab:shape} is the proposed design with \texttt{CNN}-based temporal encoder (i.e., Figure~\ref{fig:model_overall}).

\subsubsection{Comparison with Alternative Methods} \hfill\\
The last thing we want to confirm with our model design is how does it compare to commonly seen off-the-shelf machine learning solutions in production environments.
Particularly, we compare our best configuration with ready-to-deploy models like a linear model, $k$ nearest neighbor, random forest, and gradient boosting.
The result is shown in Table~\ref{tab:alternative}.
The hyper-parameters for the alternative methods are determined using 5-fold cross-validation with training data.
The percent improvement is computed against the best alternative based on each performance measurement.

\begin{table}[ht]
\centering
\caption{The proposed method outperforms common off-the-shelf machine learning models in production environment.}
\label{tab:alternative}
\resizebox{0.70\columnwidth}{!}{
\begin{tabular}{l|ccc}
Method               & RMSE   & NRMSE  & $R^2$  \\ \hline
Linear Model         & 9.1732 & 1.1039 & 0.0312 \\
$k$ Nearest Neighbor & 6.2690 & 0.8452 & 0.2217 \\
Random Forest        & 6.0354 & 0.8224 & 0.2172 \\
Gradient Boosting    & 7.7524 & 0.8670 & 0.1638 \\
\textbf{Proposed}    & \textbf{5.7147} & \textbf{0.7386} & \textbf{0.2771} \\ \hline
\% Improved           & 4.4\%  & 10.2\% & 27.3\%
\end{tabular}
}
\end{table}

Out of the four rival methods, the two more competitive ones are the $k$ nearest neighbor ($k$NN) and random forest (RF), where the $k$NN has the best $R^2$ while RF has the best RMSE and NRMSE.
When comparing the proposed model's $R^2$ with $k$NN's $R^2$, we see a $27.3\%$ improvement.
When comparing the proposed model's RMSE and NRMSE with RF's RMSE and NRMSE, we see a $4.4\%$ and $10.2\%$ improvement, respectively.
The proposed model outperforms the rivals more in NRMSE and $R^2$ based on the percent improvement.
As both NRMSE and $R^2$ are computed in a ``normalized'' scale and measure how the estimated trend matches the ground truth, the proposed model captures more details in the transaction metric sequence than alternative methods.
Note, the base models presented in Table~\ref{tab:base} are the baseline deep learning-based models.

\subsection{Evaluation of Online Learning Schemes}
\label{sec:eval_online}
To further improve the system's performance, we use the online learning framework presented in Section~\ref{sec:learning} to update the model as the system receiving previously unavailable training data.
Each day, the system receives the ground truth of the date 90 days before the current day.
The system then uses the newly received data and the existing training data to update the model.
The number of iteration~$n_{\mathrm{iter}}$ used in Algorithm~\ref{alg-train} is~100, and the batch size is~1,024.

\subsubsection{Is Online Learning Necessary?} \hfill\\
Before we compare different online learning methods, we first confirm whether performing online learning to deployed model is the right decision.
We compare the offline learning model with the naive online learning model where the model is updated with Algorithm~\ref{alg-train} using baseline sampling method (i.e., uniform sampling).
The experiment result is summarized in Table~\ref{tab:online_base}.

\begin{table}[ht]
\centering
\caption{Baseline online learning model outperforms the offline learning model.}
\label{tab:online_base}
\resizebox{0.70\columnwidth}{!}{
\begin{tabular}{l|ccc}
Method          & RMSE   & NRMSE  & $R^2$ \\ \hline
Offline         & 5.7147 & 0.7386 & 0.2771 \\
\textbf{Online Baseline} & \textbf{5.5240} & \textbf{0.7159} & \textbf{0.3258} \\ \hline
\% Improved      & 3.3\%  & 3.1\%  & 17.6\%
\end{tabular}
}
\end{table}

Online learning technique does improve the performance noticeably, especially the $R^2$.
To provide further motivation for refining the online learning method before showcasing the benchmark experiment, we present the performance gain of our best performing online learning method (i.e., segmentation + similarity) comparing to the baseline online learning method in Table~\ref{tab:online_best}.
The best performing online learning method identified by our benchmark experiment is capable of further enhancing the system.

\begin{table}[ht]
\centering
\caption{Using the best online learning model the performance can be further improved.}
\label{tab:online_best}
\resizebox{0.65\columnwidth}{!}{
\begin{tabular}{l|ccc}
Method          & RMSE   & NRMSE  & $R^2$ \\ \hline
Online Baseline & 5.5240 & 0.7159 & 0.3258 \\
\textbf{Online Best} & \textbf{5.3800} & \textbf{0.7000} & \textbf{0.3400} \\ \hline
\% Improved      & 2.6\%  & 2.2\%  & 4.4\%
\end{tabular}
}
\end{table}

\subsubsection{What is the Best Online Learning Scheme?} \hfill\\
To determine the best online learning scheme out of all the schemes outlined in Section~\ref{sec:learning} for our system, we have performed a benchmark experiment with our dataset.
For the fixed window method, we test it with a window size of 90 days (i.e., one season) and 365 days (i.e., one year).
In addition to testing the listed methods alone, we also test every combination between a temporal-based and non-temporal-based methods.
To combine the probabilities from two methods, we multiply the sampling probability of one method with the other method's sampling probability.
We test a total of 40 different online learning schemes on six different model architectures from Section~\ref{sec:verification}.
For each model, we compute each online learning scheme's rank based on either the RMSE, NRMSE, or $R^2$; then, for each online learning scheme, we compute the average rank across the six different model architectures.

\definecolor{blue}{HTML}{67A9CF}
\definecolor{orange}{HTML}{EF8A62}

The result is presented in Table~\ref{tab:online_rmse} and Table~\ref{tab:online_nrmse}.
Because the presented number is average rank, the tables' values are the lower, the better.
We also compute the mean of average rank for each row and column to help interpret the results.
Each cell is color based on its average rank relative to the baseline (i.e., \textit{Uniform+Uniform}).
If the result in a cell is better than the baseline, the cell is colored with \textcolor{blue}{blue}.
If the result in a cell is worse than the baseline, the cell is colored with \textcolor{orange}{orange}.

\begin{table}[ht]
\centering
\caption{Averaged ranking of each online learning scheme based on RMSE.}
\label{tab:online_rmse}
\resizebox{0.99\columnwidth}{!}{
\setlength{\arrayrulewidth}{1pt}
\begin{tabular}{l|ccccc|c}
RMSE             & Uniform                       & Fix (90)                      & Fix (365)                     & Decay                         & Segment                       & Average                        \\ \hline
Uniform          & \cellcolor[HTML]{FFFFFF}21.00    & \cellcolor[HTML]{FEF5F2}22.44 & \cellcolor[HTML]{C4DDEC}15.67 & \cellcolor[HTML]{E1EEF5}18.33 & \cellcolor[HTML]{C9E0ED}16.11 & \cellcolor[HTML]{FFFFFF}18.71    \\
Similar          & \cellcolor[HTML]{CAE1EE}16.22 & \cellcolor[HTML]{EFF5F9}19.56 & \cellcolor[HTML]{B3D4E7}14.11 & \cellcolor[HTML]{9BC6DF}12.00 & \cellcolor[HTML]{D2E6F1}17.00 & \cellcolor[HTML]{CFE3EF}15.78 \\
High-error       & \cellcolor[HTML]{F9CEBD}28.00 & \cellcolor[HTML]{F4AE92}32.44 & \cellcolor[HTML]{F8C7B3}29.00 & \cellcolor[HTML]{F9CEBD}28.00 & \cellcolor[HTML]{F7BEA7}30.22 & \cellcolor[HTML]{F4AC90}29.53 \\
Low-error        & \cellcolor[HTML]{9CC7E0}12.11 & \cellcolor[HTML]{84B9D8}9.89  & \cellcolor[HTML]{74B0D3}8.44  & \cellcolor[HTML]{7DB5D5}9.22  & \cellcolor[HTML]{67A9CF}7.22  & \cellcolor[HTML]{67A9CF}9.38  \\
High-confidence  & \cellcolor[HTML]{B9D7E8}14.67 & \cellcolor[HTML]{D4E6F1}17.11 & \cellcolor[HTML]{A4CBE2}12.78 & \cellcolor[HTML]{98C4DE}11.67 & \cellcolor[HTML]{98C4DE}11.67 & \cellcolor[HTML]{ABCFE4}13.58 \\
Low-confidence   & \cellcolor[HTML]{F3A789}33.44 & \cellcolor[HTML]{EF8A62}37.44 & \cellcolor[HTML]{F4AE92}32.44 & \cellcolor[HTML]{F5B094}32.22 & \cellcolor[HTML]{F3A383}34.00 & \cellcolor[HTML]{EF8A62}33.91 \\
High-variability & \cellcolor[HTML]{FBE0D5}25.44 & \cellcolor[HTML]{F6B89F}31.11 & \cellcolor[HTML]{FBDCD0}26.00 & \cellcolor[HTML]{FDEDE7}23.56 & \cellcolor[HTML]{FAD4C5}27.11 & \cellcolor[HTML]{F7C2AE}26.64 \\
Low-variability  & \cellcolor[HTML]{F7FAFC}20.33 & \cellcolor[HTML]{F2F8FB}19.89 & \cellcolor[HTML]{BBD8E9}14.89 & \cellcolor[HTML]{B3D4E7}14.11 & \cellcolor[HTML]{A7CDE3}13.11 & \cellcolor[HTML]{DAEAF3}16.47 \\ \hline
Average          & \cellcolor[HTML]{FFFFFF}21.40    & \cellcolor[HTML]{EF8A62}23.74 & \cellcolor[HTML]{84B9D8}19.17 & \cellcolor[HTML]{67A9CF}18.64 & \cellcolor[HTML]{99C5DE}19.56 &
\end{tabular}
}
\end{table}

Table~\ref{tab:online_rmse} presented the RMSE-based average rank of different online learning schemes.
When considering different temporal-based methods, \textit{Fix (365)}, \textit{Decay}, and \textit{Segment} constantly outperform the \textit{Uniform} baseline.
Generally, to learn a better model with lower RMSE, the model needs to see data more than 90 days old as 90 days fixed window performs worse than the baseline.
For non-temporal-based methods, \textit{Similar}, \textit{Low-error}, \textit{High-confidence}, and \textit{Low-variability} outperform the baseline.
Because \textit{Low-error}, \textit{High-confidence}, and \textit{Low-variability} push the model to focus on ``easy'' or ``consistent'' examples in the training data, the improvement is likely caused by the removal of noisy training examples for our data set.
Overall, combining \textit{Low-Error} with \textit{Segment} gives the best result when the performance is measured with RMSE.

\begin{table}[ht]
\centering
\caption{Averaged ranking of each online learning scheme based on NRMSE.}
\label{tab:online_nrmse}
\resizebox{0.99\columnwidth}{!}{
\setlength{\arrayrulewidth}{1pt}
\begin{tabular}{l|ccccc|c}
NRMSE             & Uniform                       & Fix (90)                      & Fix (365)                     & Decay                         & Segment                       & Average                        \\ \hline
Uniform          & \cellcolor[HTML]{FFFFFF}29.11    & \cellcolor[HTML]{9BC6DF}11.78 & \cellcolor[HTML]{CAE1EE}20.00 & \cellcolor[HTML]{FEF3EE}30.00 & \cellcolor[HTML]{81B7D7}7.11  & \cellcolor[HTML]{FFFFFF}19.60    \\
Similar          & \cellcolor[HTML]{B5D5E7}16.22 & \cellcolor[HTML]{81B8D7}7.22  & \cellcolor[HTML]{8FC0DB}9.67  & \cellcolor[HTML]{82B8D7}7.33  & \cellcolor[HTML]{67A9CF}2.56  & \cellcolor[HTML]{67A9CF}8.60  \\
High-error       & \cellcolor[HTML]{F4A98B}35.11 & \cellcolor[HTML]{E5F0F6}24.67 & \cellcolor[HTML]{FEF3EE}30.00 & \cellcolor[HTML]{EAF3F8}25.56 & \cellcolor[HTML]{BDD9EA}17.67 & \cellcolor[HTML]{F6BAA3}26.60 \\
Low-error        & \cellcolor[HTML]{F8FBFC}28.00 & \cellcolor[HTML]{90C0DC}9.78  & \cellcolor[HTML]{C4DDEC}18.89 & \cellcolor[HTML]{B2D3E6}15.67 & \cellcolor[HTML]{7DB5D6}6.56  & \cellcolor[HTML]{CAE1EE}15.78 \\
High-confidence  & \cellcolor[HTML]{FAFCFD}28.33 & \cellcolor[HTML]{A4CBE2}13.33 & \cellcolor[HTML]{D5E7F1}21.78 & \cellcolor[HTML]{B0D2E6}15.44 & \cellcolor[HTML]{7CB5D5}6.33  & \cellcolor[HTML]{DBEBF3}17.04 \\
Low-confidence   & \cellcolor[HTML]{EF8A62}37.22 & \cellcolor[HTML]{F5AF94}34.67 & \cellcolor[HTML]{F6B9A1}34.00 & \cellcolor[HTML]{F9FCFD}28.22 & \cellcolor[HTML]{DDEBF4}23.22 & \cellcolor[HTML]{EF8A62}31.47 \\
High-variability & \cellcolor[HTML]{F8CBB9}32.78 & \cellcolor[HTML]{F4F8FB}27.22 & \cellcolor[HTML]{F8FBFC}28.00 & \cellcolor[HTML]{E8F2F7}25.22 & \cellcolor[HTML]{C6DEED}19.22 & \cellcolor[HTML]{F6BCA4}26.49 \\
Low-variability  & \cellcolor[HTML]{FEF7F5}29.67 & \cellcolor[HTML]{A3CBE2}13.11 & \cellcolor[HTML]{DBEAF3}22.89 & \cellcolor[HTML]{BCD9EA}17.56 & \cellcolor[HTML]{8BBDDA}8.89  & \cellcolor[HTML]{EEF5F9}18.42 \\ \hline
Average          & \cellcolor[HTML]{FFFFFF}29.56    & \cellcolor[HTML]{9BC6DF}17.72 & \cellcolor[HTML]{C9E0EE}23.15 & \cellcolor[HTML]{B4D4E7}20.63 & \cellcolor[HTML]{67A9CF}11.45 &
\end{tabular}
}
\end{table}

According to NRMSE-based average rank (Table~\ref{tab:online_nrmse}), for temporal-based methods, \textit{Segment} gives superior performance comparing to others.
When combining \textit{Segment} with different non-temporal-based methods, the conclusion is similar to Table~\ref{tab:online_rmse}, the better methods are \textit{Similar}, \textit{Low-error}, \textit{High-confidence}, and \textit{Low-variability}.
Overall, combining \textit{Similar} with \textit{Segment} gives the best result when the performance is measured with NRMSE.
It is almost always the best method for each tested model architecture.
We also compute average rank with $R^2$, because the conclusion is similar to Table~\ref{tab:online_nrmse} we omit the table for brevity.

In conclusion, both \textit{Similar+Segment} and \textit{Low-Error+Segment} is a good choice as both have great average ranks in all performance measurements.
On one hand, \textit{Similar+Segment} has top performances when consider both NRMSE and $R^2$; therefore, we should pick \textit{Similar+Segment} if NRMSE and $R^2$ are more important.
On the other hand, although \textit{Low-Error+Segment} does not have outstanding NRMSE and $R^2$, it has the best RMSE-based average rank.
\textit{Low-Error+Segment} should be used when RMSE is more critical for the application.
Since we care more about capturing the trend rather than the raw values for our system, we choose to deliver the model trained with \textit{Similar+Segment} as prototype.
The delivered model's performance is shown in the second row of Table~\ref{tab:online_best}.
\section{Related Work}
We focus our review on research works studying \textit{online learning} and \textit{time series perdition} problems. The reasons are: 1) the proposed transaction metric estimation system utilizes online learning techniques, and 2) time series perdition models inspire the proposed model architecture design.

Online learning is a supervised learning scenario where the training data are made available incrementally in a streaming fashion. A common challenge when developing a system under such a scenario is handling a phenomenon called \textit{concept drift}~\cite{schlimmer1986incremental,gama2014survey}.
Over the years, several survey/benchmark papers of various scopes have been written on this topic~\cite{vzliobaite2010learning,gama2014survey,krawczyk2017ensemble,hoi2018online,losing2018incremental,lu2018learning}.
There are many facets to the development of an online learning solution for the concept drift problem.
For example, detection of concept drift is a popular sub-problem studied in~\cite{bifet2007learning,liu2017regional,yu2017concept}.
There are also online learning solutions focusing on adapting a specific machine learning model to the concept drift like~\cite{platt1998sequential} for support vector machine,~\cite{oza2005online} for boosting algorithm, and~\cite{losing2016knn} for $k$ nearest-neighbor model.
Developing application-specific solutions like~\cite{jia2017incremental}, which is designed for land cover prediction, is another common research direction.
This work focuses on developing an online learning method specific for a deep learning-based transaction metric estimation system.

Time series prediction or forecasting is a research area with many different problem settings~\cite{de200625,box2015time,faloutsos2019forecasting}.
The proposed transaction metric estimation system is mainly focused on a variant of the problem called multi-horizon time series prediction~\cite{taieb2015bias}.
There have been many models proposed for the multi-horizon time series prediction problem~\cite{taieb2015bias,wen2017multi,shih2019temporal,fan2019multi,zhuang2020multi,yeh2020multi}.
Taieb and Atiya~\cite{taieb2015bias} test multiple multi-horizon time series prediction strategies and show making multi-horizon prediction by merely making a 1-step recursive prediction is not always the best strategy.
Wen et al.~\cite{wen2017multi} combine the direct strategy from~\cite{taieb2015bias} with various recurrent neural network-based model architecture for multi-horizon time series prediction.
Zhuang et al.~\cite{zhuang2020multi} further refine the recurrent neural network-based model to capture the hierarchical temporal structure within the input time series data.
Aside from recurrent neural network-based model, Fan et al.~\cite{fan2019multi} examine the possibility of adopting an attention mechanism for multi-horizon time series prediction.
Shih et al.~\cite{shih2019temporal} propose a network structure that combines ideas from the attention mechanism and convolution neural network.
Sen et al.~\cite{sen2019think} combine convolutional neural network-based design and matrix factorization model to capture the time series's local and global properties at the same time.
Based on similar principles,~\cite{yeh2020multi} propose a convolutional neural network-based network design with sub-networks dedicated to capturing the ``shape'' and ``scale'' information of time series data independently.
The model used in the proposed transaction metric estimation system is inspired by works introduced in this paragraph with an extension to handle the multimodal nature of our transaction data.
\vspace{-1em}
\section{Conclusion and Future Work}
We propose an online transaction metric estimation system that consists of a user-facing interface, an online adaptation module, and an offline training module.
We present a novel model design and demonstrate its superb performance compared with the alternatives.
Further, we show the benefit of applying online learning schemes to the problem.
To identify the best online learning scheme for our system, we perform a detailed set of experiments comparing existing and novel online learning schemes.
The proposed system not only been used as a monitoring tool, but also provides the predicted metrics as features to our in-house fraud detection system.
For future work, we are looking into applying the findings to other entities from payment network like merchants, issuers, or acquirers.
We are also interested in exploring the possibility of test the system on data from other domains.

\bibliographystyle{ACM-Reference-Format}


\end{document}